# An Appearance Defect Detection Method for Cigarettes Based on C-CenterNet

Hongyu Liu, Guowu Yuan*, Lei Yang, Kunxiao Liu, and Hao Zhou

School of Information Science and Engineering, Yunnan University, Kunming 650504, China; liuhy@mail.ynu.edu.cn
* Correspondence: gwyuan@ynu.edu.cn;

**Abstract:** Due to the poor adaptability of traditional methods in the cigarette detection task on the automatic cigarette production line, it is difficult to accurately identify whether a cigarette has defects and the types of defects; thus, a cigarette appearance defect detection method based on C-CenterNet is proposed. This detector uses keypoint estimation to locate center points and regresses all other defect properties. Firstly, Resnet50 is used as the backbone feature extraction network, and the convolutional block attention mechanism (CBAM) is introduced to enhance the network's ability to extract effective features and reduce the interference of non-target information. At the same time, the feature pyramid network is used to enhance the feature extraction of each layer. Then, deformable convolution is used to replace part of the common convolution to enhance the learning ability of different shape defects. Finally, the activation function ACON (ActivateOrNot) is used instead of the ReLU activation function, and the activation operation of some neurons is adaptively selected to improve the detection accuracy of the network. The experimental results are mainly acquired via the mean Average Precision (mAP). The experimental results show that the mAP of the C-CenterNet model applied in the cigarette appearance defect detection task is 95.01%. Compared with the original CenterNet model, the model's success rate is increased by 6.14%, so it can meet the requirements of precision and adaptability in cigarette detection tasks on the automatic cigarette production line.

## 1. Introduction

Cigarettes are the main product in the tobacco industry. The appearance quality of cigarettes directly reflects the production level of cigarette factories. However, in the final step of making cigarettes from tobacco, during the packaging process of cigarettes, due to the limitations of the production process, various defects will inevitably appear on the surfaces of cigarettes, and these defects affect the appearance and brand image of cigarette products. The industry places a strong emphasis on quality, strictly controls the quality of cigarettes, and does not allow cigarettes with defects in appearance to enter the market. The traditional method of identifying appearance defects in cigarettes generally relies on manual labor, which is unstable and increases the costs of actual production. Therefore, the purpose of this paper is mainly to improve the detection accuracy of cigarette appearance defects, to achieve high-quality automatic detection of cigarette appearance defects, and to improve the quality of cigarette products.

Currently, most studies on the appearance of cigarettes use traditional digital image processing methods, which usually include feature extraction and image segmentation. Qu H et al. performed image smoothing, edge detection, binarization and feature extraction on the cross-sectional image of the filter rod, analyzed the region of interest, and finally obtained the number of filter rods [1]. Li M X et al. proposed the minimum outer rectangle to be applied to cigarette label defect shape analysis, and realized the defect detection of printed images [2]. Feng S et al. introduced image segmentation and morphological operations to distinguish cigarette regions from the background, and then used a trilinear model to locate each cigarette and determine whether the cigarette has defects by calculating the number of pixels in each obtained region [3]. Li C J et al. designed a self-learning control system based on a data acquisition card, and used a database, a data acquisition card and an industrial computer to detect the surface defects of cigarettes [4]. Xiao Z Y extracted the cigarettes using canny operator and then determined the defects by analyzing the area ratio of the incomplete part of the grayscale image [5]. Li J et al. used the maximum contour area determination method to detect cigarettes with significant cosmetic defects, and then used the template matching method to detect them with slight defects [6]. However, these traditional methods are less adaptable to complex situations and often require a combination with manual labor. It may lead to poor generalization of defect detection methods. In addition, the detection accuracy of this method is low, and there are many erroneous detection results. It is only effective for defects with distinct edges, simple backgrounds and relatively flat surfaces.

In recent years, deep learning technology has been widely used to solve some traditional industrial problems due to its advantages of strong learning ability and good portability. Defect detection is an aspect of target detection. Currently, deep learning-based defect detection methods commonly used include R-CNN, Faster R-CNN [7-9], SSD [10], YOLO [11-14] series, CenterNet [15], etc. In some defect detection studies similar to studies of cigarette appearance defects, a vast number of research results have been obtained. For example, Xu L et al. based on the original framework of Faster R-CNN, performed clustering optimization on anchor points, replaced the ROI pool with ROI Align, and applied it to sand inclusion defect detection [16]. Huang F R et al. proposed a Faster R-CNN-based part surface defect detection algorithm based on the cluster generation anchor scheme, and introduced a multi-level ROI pooling layer structure to achieve the efficient and accurate detection of part surface defects [17]. Xu Y et al. proposed a Path-Enhanced Feature Pyramid Network (PAFPN) and an edge detection branch integrated into the Mask R-CNN, and this was applied in tunnel defect detection and segmentation [18]. Zhang F H et al. combined Multi-Scale Overlapping Sliding Pooling (SOSP) and proposed an SSD-based jelly impurity detection method [19]. Chen S H et al. replaced the Darknet-53 backbone network with the densely connected convolutional network DenseNet and proposed an LED chip defect detection method based on the YOLOv3 network [20]. Song Y N et al. adopted the YOLOv3 algorithm framework and introduced dimensional clustering, which was applied to the detection of rail surface defects [21].

However, the research on the detection and classification of cigarette appearance defects by deep learning methods started relatively late, and only a few studies have tried deep learning methods. Although Neural networks have been proven to be effective in many detection tasks, there are still many challenges to the detection of cigarette appearance defects: the data set of cigarette appearance defects is insufficient, the defect scale changes greatly, and the cigarette image is narrow and long. Qu R et al. improved the SSD network with methods such as pyramid convolution to realize the detection of cigarette appearance defects with complex features and imbalanced datasets [22]. Li L F et al. used MobileNet to replace Vgg16 of traditional Faster R-CNN to effectively detect cigarette capsule defects [23]. Their method demonstrates the effectiveness of CNN for cigarette appearance defect detection. However, the detection recall rate and precision of the above-mentioned deep learning detection algorithms cannot meet the current actual production needs and cannot be applied to actual production.

Based on the above discussion, on the basis of previous research, this paper conducts research on related deep learning algorithms for cigarette appearance defect detection, and proposes a CenterNet-based cigarette appearance defect detection method, which is robust to cigarette appearance defect detection and has higher detection preci-sion and recall rate. Due to the good detection performance of CenterNet in many defect detection tasks [24], CenterNet is adopted as the structure of our neural network. In addition, in order to better adapt to the characteristics of cigarette images and improve the recognition effect of cigarette defects of different scales, we have made the following improvements to the network structure: (1) for the original cigarette data set, appropriate data enhancement and balancing of the data set are carried out, mainly using methods such as flipping, cropping, brightness adjustment, adding noise, synthesizing new samples, and generating adversarial networks, to enrich the data and enhance the generalization of the network; (2) we add the Convolution Block Attention Mechanism (CBAM) to the backbone feature extraction network, which improves the sensitivity to the detection target; (3) we introduce the Feature Pyramid Network (FPN) to integrate a multi-scale feature map; (4) we replace the last standard convolution block of the backbone detection network with deformable convolution, which improves the recognition effect of irregular cigarette defect targets; (5) we optimize the activation function, replace ReLU with the ACON activation function, and adjust the activation operation of neurons adaptively, which is beneficial to improving the adaptive ability of the network and improving the detection accuracy of the network. The experimental results show that the improved model has improved precision, recall and average detection accuracy on 5 common cigarette appearance types data sets, and at the same time, the average detection speed can also meet most of the detection requirements.

**2. Materials and Methods**

*2.1 CenterNet*

C-CenterNet takes CenterNet as the baseline. CenterNet is a one-stage detection method, and its main feature is that it does not detect objects based on an a priori box. This method models an object as a single point, i.e., the center point of its bounding box, through a heat map, obtaining the center and then returning other information about the object, such as width, height, position, etc. The overall idea is relatively simple. The network eliminates the need for complex a priori frame design, reduces the parameter selection, and eliminates the non-maximum suppression post-processing process. This makes the network less computationally intensive. In addition, CenterNet only uses 4x down-sampled high-resolution feature maps, which is a higher resolution than the 16x image-scale scaling performed in most object detection algorithms, making it suitable for small objects.

The CenterNet model uses three networks—Hourglass [25], DLA [26], and Resnet [27]—as the backbone network, and the three networks are complete encoding–decoding networks. The CenterNet detection algorithm framework based on Resnet50 is shown in Figure 1 below. For the input image, first, the down-sampling feature map is output after encoding by the CenterNet backbone network; then, the resolution of the output feature map is improved by up-sampling by the decoder, and it is finally detected by the detection layer. The output feature map of the detection layer to the backbone network undergoes a series of convolution operations, including a regression heat map, center point bias map, and target box size prediction map, and outputs the final prediction result.

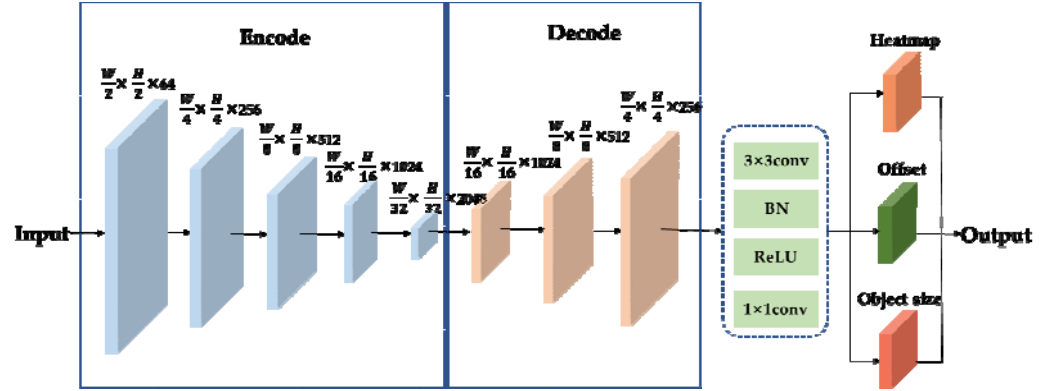

**Figure 1**. Overall structure of CenterNet based on Resnet50

*2.2 Loss function*

The loss function of CenterNet includes the heat map loss $L_{cls}$, center point bias loss $L_{off}$, and regression width and height loss $L_{reg}$ [15]. The calculation of the heat map loss adopts focal loss.

$$L_{cls} = -\frac{1}{N}\begin{cases}(1-\hat{Y}_{xyc})\log(\hat{Y}_{xyc}) & Y_{xyc}=1 \\ (1-Y_{xyc})^{\beta}(\hat{Y}_{xyc})^{a}\log(1-\hat{Y}_{xyc}) & Y_{xyc}\neq 1\end{cases} \quad (1)$$

Specifically, $N$ is the number of targets, $Y_{xyc}$ is the actual pixels, $\hat{Y}_{xyc}$ is the predicted pixels, and the hyperparameters $\alpha=2$, $\beta=4$.

Since the heat map is obtained by down-sampling, the center point has a certain deviation from the original feature map, so it is necessary to calculate the center point offset loss, using L1 Loss, as follows:

$$L_{off} = \frac{1}{N}\sum_{p}\left|\hat{O}_p - \left(\frac{p}{R}-\bar{p}\right)\right| \quad (2)$$

Among the terms, $\hat{O}_p$ is the prediction center point bias, $R$ is the down-sampling multiple, $\frac{p}{R}$ is the center point subsampling coordinates, and $\bar{p}$ is the center point coordinates obtained after taking the whole down $\frac{p}{R}$.

Similarly, the regression width and height loss also use L1 Loss, which is calculated as follows:

$$L_{reg} = \frac{1}{N}\sum_{k=1}^{N}\left|S_{F_i} - S_k\right| \quad (3)$$

where $S_{F_i}$ is the network-predicted width and height loss, and $S_k$ is the regression of the width and height.

The linear combination of the three is the total loss function:

$$L_{dot} = L_{cls} + \lambda_{reg}L_{reg} + \lambda_{off}L_{off} \quad (4)$$

Among the terms, the coefficient $\lambda_{reg}=0.1$, $\lambda_{off}=1$.

*2.3 C-CenterNet for the detection of cigarette appearance defects*

2.3.1 C-CenterNet

The model mainly consists of two parts. The first part is the backbone network, and the main task of this part is to extract image features and generate feature maps. CenterNet provides three backbone network schemes, but due to the small data sets in this paper, if the DLA or Hourglass methods are used, the data sets are not sufficient and may easily lead to overfitting, and the two methods have many parameters, complex implementations, and are inconvenient for practical applications. Therefore, the backbone network used in this paper is Resnet50. According to the characteristics of the cigarette appearance image dataset, in the Resnet50 network part, we introduced FPN for feature enhancement and fusion to improve the feature extraction effect for small targets in the cigarette data set. Moreover, the basic residual structure has also been improved: (1) an attention mechanism is introduced to suppress the interference of irrelevant information, so that feature extraction is more focused on the target itself; (2) we optimize the activation function, replace part of the ReLU function with the ACON activation function, and adaptively choose whether to activate and in what way to activate neurons; (3) we optimize the convolution structure and replace the partial convolution structure at the back with a deformable convolution structure. Compared with traditional convolution, this system can capture more irregular-shaped feature information, making it suitable for the defect-type targets described in the paper and able to better solve the problem of large differences in shape in a cigarette data set.

The second part is the prediction of the results. In this part, the network regards the complex object detection problem as a simple center point detection problem. The input feature map obtains the heat map, in which the position of the central point can be calculated from the heat map, and then the center point position is adjusted according to the offset to predict the box width and height to obtain the final detection result. The proposed structure of the cigarette appearance defect detection network for C-CenterNet is shown in Figure 2.

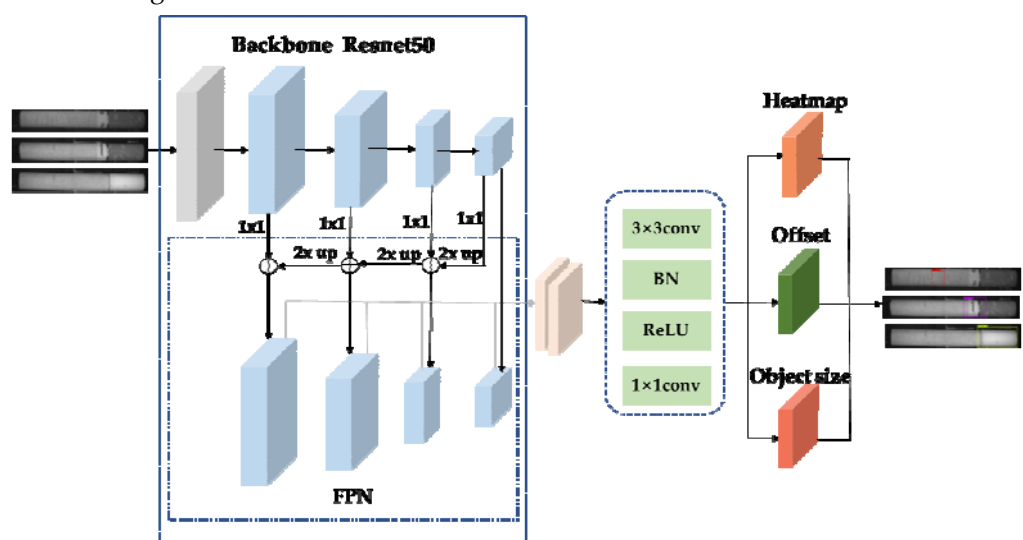

**Figure 2.** Overall structure of C-CenterNet

2.3.2 Backbone feature extraction network

The backbone feature extraction network is mainly based on the Resnet50 framework. In the backbone feature extraction network, we first perform one-layer convolution, BN batch normalization, the ReLU activation function, and maximum pooling; then we complete the feature information extraction through four residual structure blocks. Each residual structure block is composed of a Conv Block and Identity Block, and the number of Identify Block stacks of the four residual blocks is different: 2, 3, 5, and 2, respectively. Subsequently, the FPN method integrates the extracted feature layers and feature fusion to obtain a high-resolution feature map.

To improve the adaptability of the backbone network on the data set, the hopping structure of the residual block of the backbone feature extraction network was improved, as shown in Figure 3. First, the attention mechanism was added after the convolution to improve the detection accuracy of the network on the cigarette appearance data set. In addition, the ReLU activation function after the second convolution of the residual module was replaced by the ACON activation function, which adaptively selects the accuracy of the cigarette appearance defect detection while introducing as few parameters as possible.

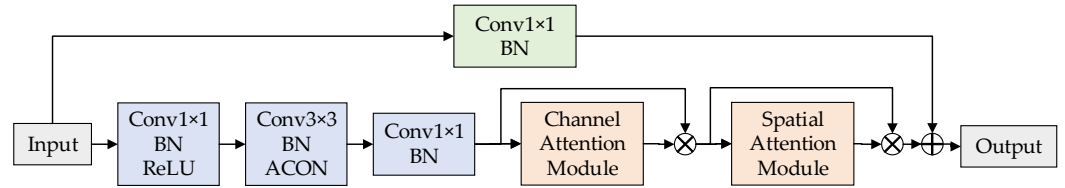

**Figure 3.** The first three layers of the residual module diagram

In particular, this paper also replaces the 3×3 convolution in the residual block of the last layer with a deformable convolution, enhancing the detection accuracy of irregular defect targets with sufficient feature learning. The schematic diagram of the structure of the last layer of the residual block is shown in Figure 4.

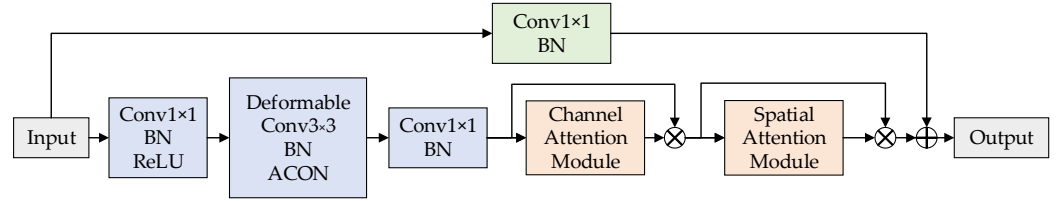

**Figure 4.** The last-layer residual module diagram

2.3.3 Strengthening of the feature extraction module

In the process of cigarette appearance defect detection, there are few defect information elements for small targets in the image, such as only simple down-sampling and up-sampling combination, and low-level information elements can easily cause loss. Therefore, in this paper, we adopt the Feature Pyramid Network (FPN) [28] to enhance the feature extraction ability, strengthen the fusion of feature information at different scales and different layers, and reduce the computational cost.

As shown in Figure 5, the FPN method first extracts features from the original image, gradually reduces the feature map resolution through convolution and pooling, and integrates features on feature maps C2, C3, C4, and C5. First, it uses P5 by convoluting C5 and transforms C4 into the same number of channels as P5; P5 samples the same feature map size as C4 by bilinear interpolation, and adds the two feature maps with pixels to obtain P4, as shown in the dashed box of Figure 5. By analogy, according to this method, a top-down path is formed, obtaining the final feature map for the prediction.

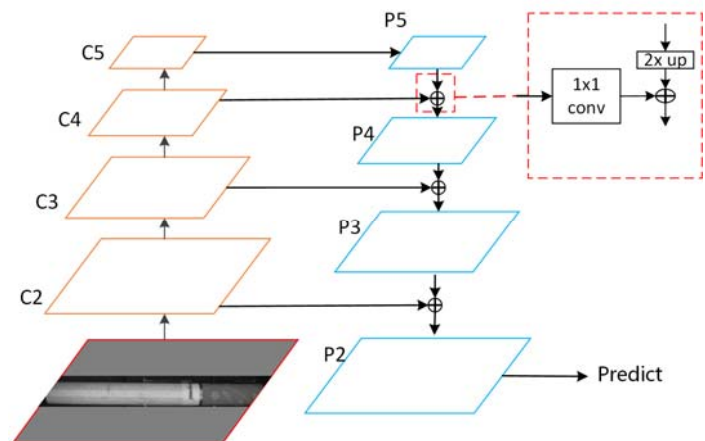

**Figure 5**. Feature pyramid structure

In this paper, the output feature maps of the second to fifth layers of the backbone feature extraction network Resnet50, i.e., C2, C3, C4, and C5, are taken as input to the FPN. Here, we perform 1×1 convolution of C5, C4, and C3 to C2 and bilinear interpolation fusion, namely layer-by-layer with the bilinear interpolation method, with the size of the previous layer and convolved feature map, which are to form a top-down path; however, we only use the feature diagram generated in the last step to generate the detection box. The introduction of FPN can effectively integrate the semantic and feature information of high and low layers and improve the accuracy of the detection of cigarette appearance defects, especially some small-sized defects.

2.3.4 Convolutional block attention mechanism

Due to the large number of network structure layers, it is easy to introduce some invalid feature information during the repeated sampling and fusion, which reduces the attention power of the model to the target features. Therefore, this paper combines the attention mechanism module of space and channel (Convolutional Block Attention Module, CBAM) [29] in the backbone feature extraction network. The addition of channel attention enables the network to pay better attention to the feature information in each channel, and to automatically obtain the importance of the existence of each piece of feature channel information through learning. Adding spatial attention enables the network to pay better attention to the information about the location. By establishing the internal relationship among regions with useful information, we can analyze and compare which regions contain useful information and which regions contain less important information. Given the characteristics of the defect target size of the image and most of the defect targets, introducing an attention mechanism can help to learn the target features in the cigarette appearance image, reduce the interference of background features, and emphasize the target information, so as to improve the detection accuracy.

In this paper, the CBAM module is added to the residual structure of Resnet50, which uses few parameters and does not affect the real-time performance of the model. The model with the added CBAM module has better performance and better interpretability than the benchmark model and focuses more on the target object itself. The Sigmoid activation function of CBAM was also replaced with the h-Sigmoid [30] activation function, avoiding the complexity of the exponential computation and reducing the computation time, and the experiments show that this substitution achieves less loss and better accuracy.

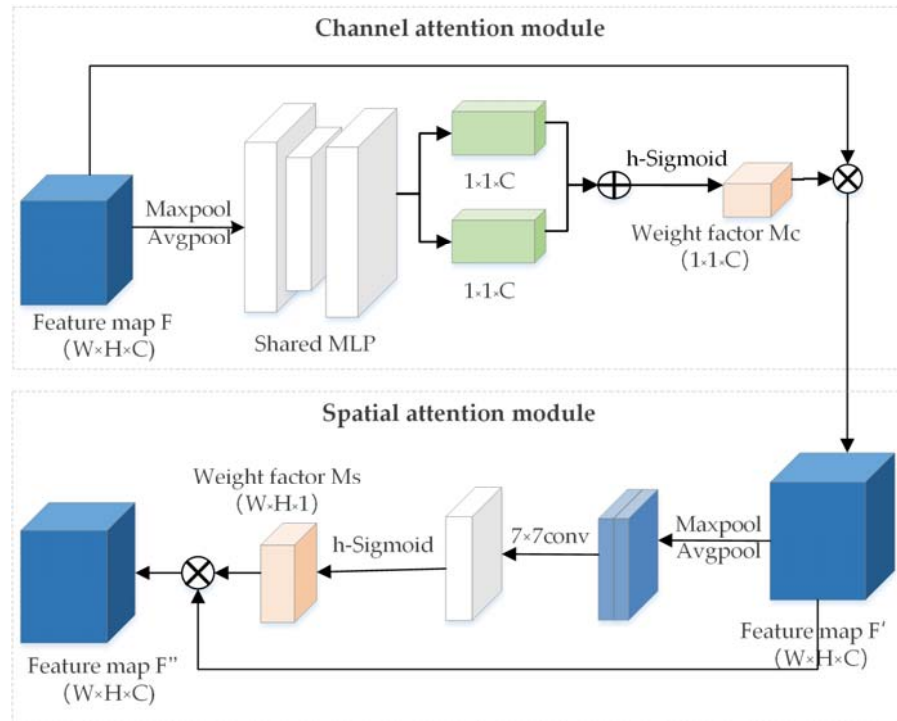

Figure 6. The CBAM module structure presented in this paper

2.3.5 ACON activation function

The activation function of the model is mainly ReLU, but many studies show that not all cases require the activation of neurons; the Swish activation function [31] has been used in some cases, which can achieve better results than ReLU. The ACON (Activate or Not) activation function [32] was proposed by Ningning Ma et al., and it is mainly characterized by the ability to adaptively learn to activate or not activate neurons. The paper treats the Swish activation function as an approximation of ReLU smoothing, making Swish a special case of ACON. The ACON-C activation function is used, and the calculation formula is as follows:

$$f_{ACON-C}(x) = S_\beta(p_1 x, p_2 x) = (p_1 - p_2)x \times \sigma[\beta(p_1 - p_2)x] + p_2 x \tag{5}$$

Among the terms, parameter $\beta$ controls the smoothness $S_\beta$, and parameter $p_1$、$p_2 (p_1 \neq p_2)$ can be learned to control the upper and lower bounds of the function. Through different values of $\beta$, $p_1$, and $p_2$, the function can be adaptively transformed into expressions of activation functions such as ReLU, Swish, and Maxout [33].

The ACON activation function unifies activation functions such as ReLU and Swish into one expression, adaptively adopting activation functions more suitable for neurons. In this paper, the advantages of the ACON activation function are used to replace the ReLU activation function after the second convolution of Bottleneck in the backbone feature extraction network with the ACON activation function. The later experimental results show that the accuracy of cigarette appearance defect detection is improved by the use of this activation function.

2.3.6 Deformable convolution

The characteristics of cigarette defects, such as malposed and folded types, are long, narrow, rich, and diverse. Ordinary convolution in Resnet50 performs better for neat and regular shapes, and poorly for targets with non-fixed shapes. Deforming convolution [34] is an improvement of the ordinary convolution, where the offset of the sampling point is introduced to obtain a flexible receptive field and realize the adaptive deformation convolution. The calculation of ordinary convolution operations is shown in (6), while the calculation formula of deformable convolution is shown in (7).

$$y(P_0) = \sum_{P_n \in R} w(P_n) x(P_0 + P_n) \tag{6}$$

Among the terms, $P_0$ is the position of a point on the feature map, $y(P_0)$ is the output feature value corresponding to $P_0$, and $R$ is the range of the receptive field area, while $P_n$ enumerates all grid points in the $R$ area, $w(P_n)$ represents the convolution kernel weight of $P_n$, and $x(.)$ represents the sampling value at the corresponding position on the feature map $x$.

$$y(P_0) = \sum_{P_n \in R} w(P_n) x(P_0 + P_n + \Delta P_n) \tag{7}$$

Among these terms, $\Delta P_n$ represents the offset to the sampling point.

As can be seen from Equations (6) and (7), the improvement of deformable convolution compared to ordinary convolution mainly lies in introducing the offset of sampling points, which realizes a convolution that can adapt to different geometries. Considering the balance of model speed and accuracy, as well as the effectiveness of the offset, only the last block in the backbone feature extraction network structure is replaced with deformable convolution, further improving the adaptability to irregular cigarette appearance defect detection when a large number of features have been learned.

## 3. Experiment and Result Analysis

### 3.1 Data Introduction

The cigarette is mainly divided into two parts: the long white part on the left is called the cigarette stick, which is used to hold the cut tobacco; the darker and shorter part on the right is called the filter tip, which is used to hold the filter cotton, as shown in Figure 7.

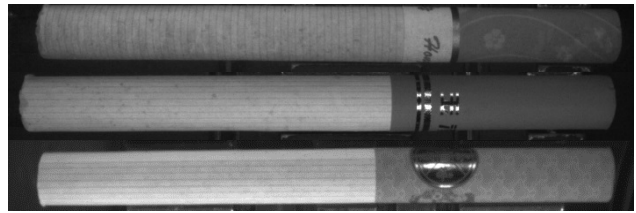

**Figure 7**. Normal cigarette image

In the production process of cigarettes, according to the causes of appearance defects in the production line, tobacco companies divide the appearance defects of cigarettes into the following four categories: dotted, folded, malposed, and unfiltered cigarettes.

Dotted cigarettes present black spots, stains, etc., of different sizes on the surfaces of the cigarettes, which are mainly formed by the unqualified printing of the cigarette stick paper or the dyeing in the later stage. Folded cigarettes mainly present some wrinkle-like shapes on the surface, which are mainly caused by the production machine. It is caused by improper operation when rolling the filter tip with the filter paper or rolling the shredded tobacco with the cigarette stick paper. Malposed cigarettes are not aligned in the process of rolling the shredded tobacco, which is mainly caused by the loosening of the packaging machine. Unfiltered cigarettes are mainly caused by the production ma-chine. The problem is caused by an inability to wrap the filter paper. Images of the specific cigarette appearance defects are shown in Figure 8.

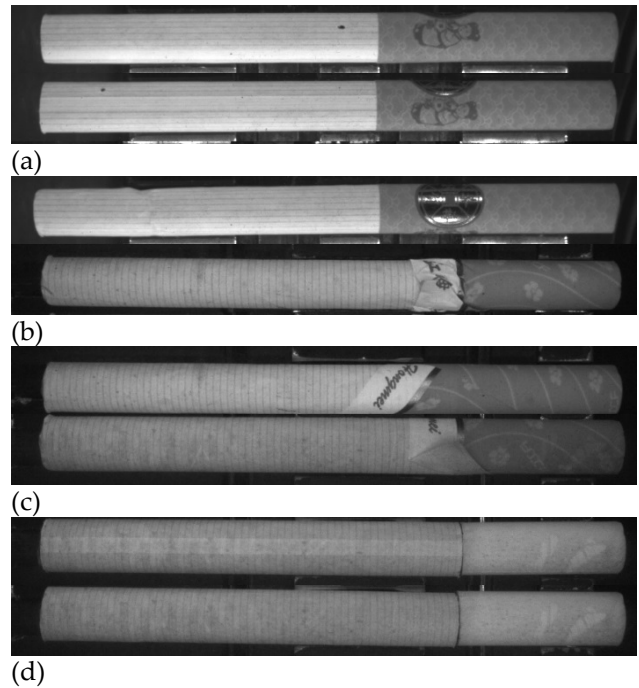

(a)

(b)

(c)

(d)

**Figure 8**. Defective cigarette types: (a) dotted cigarettes; (b) folded cigarettes; (c) malposed cigarettes; (d) unfiltered cigarettes

In the data set considered in this paper, the detection targets are divided into normal, dotted, folded, malposed, and unfiltered cigarette types. Due to the uneven quality of the pictures taken by industrial high-speed cameras, the obtained data set was screened, and only 2,000 valid pictures were obtained. Due to the limited cigarette appearance defect data set, to enable the network to learn better and achieve better generalization and adaptability, we adopted flip transformation, random tailoring, brightness transformation, the synthesis of new samples, and the generation of an adversarial network for data enhancement. The enhanced data set distribution is shown in Figure 9.

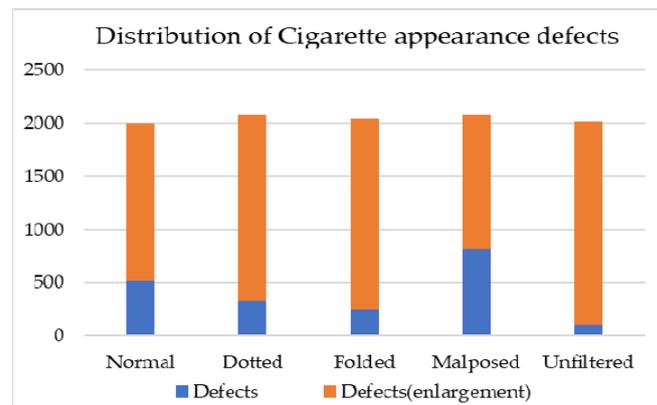

**Figure 9**. Distribution diagram of cigarette appearance data set

In order to improve the generalization ability of the model and avoid over-fitting, data enhancement is performed on the pictures, and the number of pictures is expanded to 10,000. Among them, 233 images belong to multiple defect categories. In order to avoid the large difference between the number of sample categories, which causes the model to focus on categories with a large number of samples, and "disregard" categories with a small number of samples, our data enhancement makes the distribution of different de-fects close. For all images, they were divided at a ratio of 6:2:2. In other words, 6000 images were used as the training set, 2000 images were used as the validation set, and the remaining 2000 annotated images were used as the test set. The division of the data set is shown in Table 1.

**Table 1.** The division of the data set of cigarette appearance pictures

| Type* | training set | validation set | test set |
|---|---|---|---|
| Normal | 1200 | 400 | 400 |
| Dotted | 1246 | 415 | 415 |
| Folded | 1227 | 409 | 409 |
| Malposed | 1248 | 416 | 416 |
| Unfiltered | 1213 | 404 | 404 |

* The single number of each type is 1200 for training set, 400 for validation set, and 400 for test set, and the ratio is 6:2:2. The rest are both this and other types.

In the process of training and testing, CenterNet needs to use the coordinate position of the target. Therefore, we used the tool LabelImg to mark the data after enhancement, and formed an annotation box XML file, as shown in Figure 10.

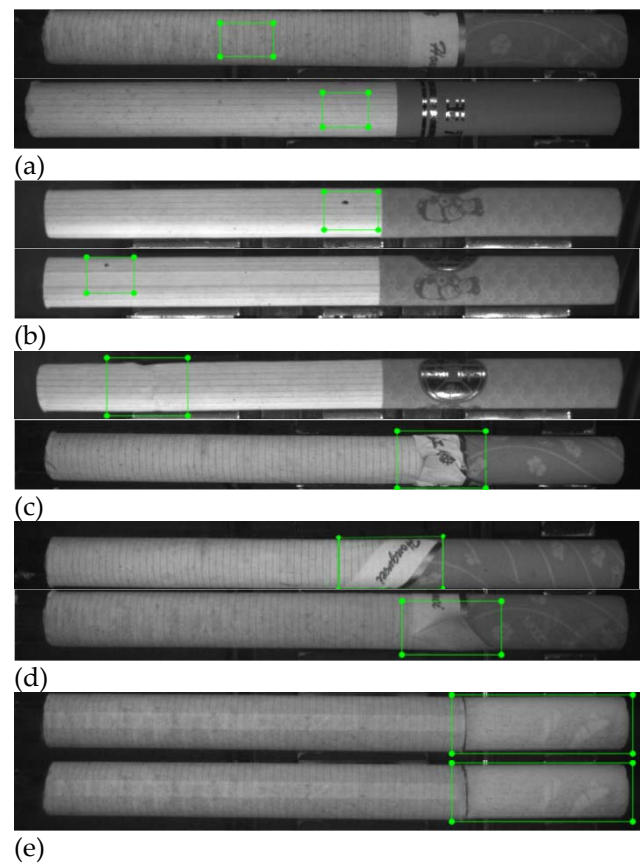

(a)

(b)

(c)

(d)

(e)

**Figure 10**. Example of cigarette labeling: (a) normal cigarette label; (b) dotted cigarette label; (c) folded cigarette label; (d) malposed cigarette label; (e) unfiltered cigarette label

This paper presents statistics on the defect aspect ratio and defect area of the enhanced and labeled cigarette appearance defect images. As shown in Figure 11 below, they were all unbalanced. The defect aspect ratio of cigarette appearance was mostly in the (1.5,7.5) interval, while the defect area ratio of cigarette appearance was mostly in the (0.05,0.075) and (0.2,0.225) intervals. Therefore, it can be seen that the paper had a large shape difference, with a small and narrow defect target.

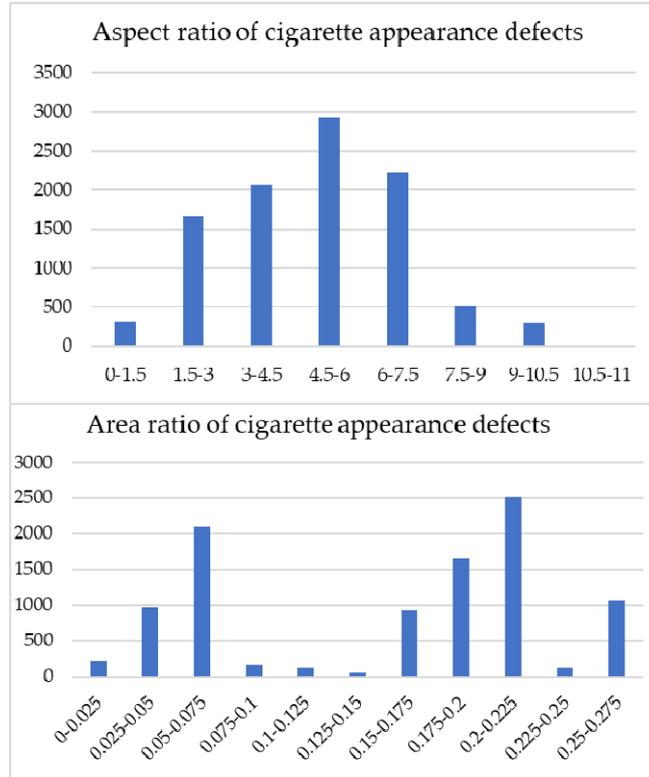

**Figure 11**. Details of cigarette appearance defect data

*3.2 Experimental parameter setting*

The experimental environment was Windows, Pytorch version 1.4.0, Cuda version 10.1, GPU was NVIDIA GTX2080Ti, and video memory was 11G.

We completed the construction of the network model according to the improved part, and trained the model for a total of 300 epochs. Among them, in order to avoid the model falling into the local optimum, the learning rate adjustment method is set to cosine annealing decay. In order to speed up the training speed and prevent the weights from being destroyed in the early stage of training, the frozen training method is adopted. The first 50 epochs are frozen for training, and each 32 images are used as a Bach Size. After 50 epochs of training, it is thawed. Each 16 images are used as a Bach Size. After each epoch is completed, the weights are updated and saved. The weight decay rate is set to 0.5. The initial learning rate is set to 0.001 when freezing, and 0.000125 after thawing. At 300 iterations, the network achieved the lowest loss value, saving the final model and input pictures for prediction.

*3.3 Training process analysis*

Figure 12 shows the change plot of the total loss values based on the C-CenterNet method and CenterNet on the cigarette appearance data set. In the first 50 epochs, the loss value of the model dropped sharply and fluctuated somewhat, showing a slow decline between 50 and 250 epochs, and the loss value gradually stabilized at 250-300 epochs and tended to converge, so 300 epochs was taken as the number of training iterations of the model. In addition, the dashed line represents the CenterNet loss function curve, and the solid line represents the C-CenterNet loss function curve; it can also be seen from the figure that, under the same conditions in which the loss function declined faster, with faster trend convergence, the final convergence value was lower than the original loss function, so, during the training process, the performance of the model was better.

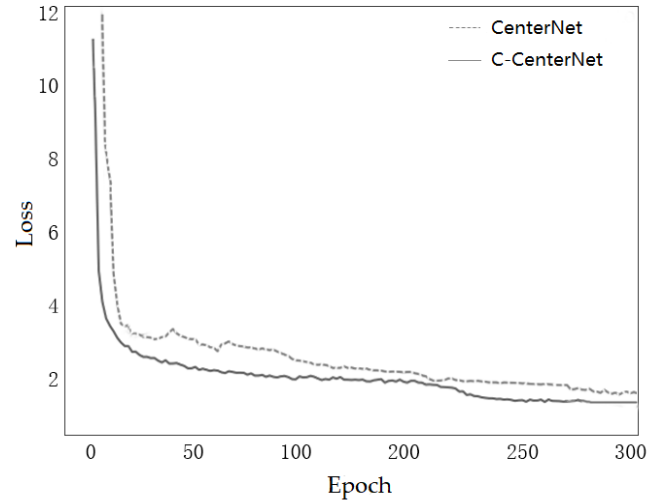

**Figure 12.** The training process loss performance

*3.4 Algorithm evaluation index*

The experiment used precision, recall, *mAP*, and *mSpeed* to evaluate the performance of the algorithm. *mSpeed* refers to the time it takes to complete a cigarette test.

The calculation formulas of precision (*P*) and recall (*R*) are as follows:

$$P = \frac{TP}{TP + FP} \quad (12)$$

$$R = \frac{TP}{TP + FN} \quad (13)$$

Among the terms, $TP$ refers to the number that is predicted to be a certain category and actually belongs to this category, and $FN$ refers to the number that is not predicted to be a certain category and actually belongs to this category. $FP$ refers to the number that is predicted to be in a category but does not actually belong in this category.

After obtaining the $P$ and $R$ of each category, a precision–recall (*P-R*) curve can be obtained, and the area enclosed by the curve and the coordinate axis is the value of $AP$, so the calculation formula of $AP$ is:

$$AP = \int_0^1 PRdR \quad (14)$$

$mAP$ seeks to calculate the average of the $AP$ values of all categories, and the calculation formula is:

$$mAP = \frac{1}{n}\sum_{i=1}^{n} AP_i \quad (15)$$

where $n$ is the number of classes divided and $AP_i$ is the $AP$ value for the i-th class.

*3.5 Contrast algorithm*

In order to verify how well the algorithm compared with other algorithms, we trained YOLOv5 and other mainstream object detection models, such as Faster R-CNN, and SSD, etc., on the dataset to compare the performance of the method proposed in this paper and other methods on the above indexes. Results of comparison experiments for the detection of the various models are shown in Table 2, where the mAP in the table refers to the mean of all AP when the IoU is set to 0.5.

**Table 2.** Comparison of the results of various algorithms

| algorithm | P /% | R /% | mAP /% | mSpeed /(ms/ branch) |
|---|---|---|---|---|
| Fast R-CNN | 80.12 | 70.02 | 71.82 | 32.3 |
| Faster R-CNN | 79.71 | 73.44 | 79.99 | 30.9 |
| SSD | 81.29 | 69.33 | 84.90 | 14.9 |
| YOLOv4 | 88.01 | 83.12 | 88.41 | 6.1 |
| YOLOv5 | 89.89 | 82.98 | 90.73 | **4.8** |
| CenterNet | 92.65 | 78.30 | 88.87 | 7.7 |
| C-CenterNet | **99.89** | **85.96** | **95.01** | 8.9 |

It can be concluded from the analysis of the results in Table 2 that, although the method proposed in this paper is not optimal in terms of average detection speed and is slower than YOLOv5, the accuracy rate, recall rate, and mAP are the highest in the table, indicating that the method yields fewer misses and is more accurate. Although the detection speed is not as good as that of YOLOv5, the detection speed achieved by the method in this paper can meet the detection needs of most current situations [35], and is much faster than the detection speed of existing research [22,23]. In addition, compared with CenterNet, the method demonstrates 7.24% higher accuracy, 7.66% higher recall, and 6.14% higher mAP, and the average detection time was increased by only 1.2ms / branch. The experimental results show that the present model is a good cigarette appearance detection method to meet the demands of cigarette target detection on the production line.

Table 3 shows the comparison of the accuracy of the data sets of the above algorithms. Due to the complex and diverse defects of the malposed type, the detection performance of Fast R-CNN, Faster R-CNN, SSD, YOLOv4, and YOLOv5 was poor. Due to its detection characteristics based on the center point, the original model of malposed types achieved better results than the other algorithms. The detection effect of C-CenterNet was better than that of CenterNet. Since the normal type has no obvious characteristics, CenterNet and Faster R-CNN performed poorly and could not distinguish their type correctly. After the improvement of CenterNet, the detection accuracy was significantly improved on normal types, exceeding the original YOLOv5 algorithm. For the detection of dotted, unfiltered, and folded types, Faster R-CNN, SSD, etc., performed generally well, while YOLOv5 was slightly better than CenterNet, and the proposed method was better than YOLOv5. Therefore, as shown in Table 3, the detection accuracy of the method on five defects and their average detection accuracy was higher than that other comparison algorithms, indicating that it has achieved an ideal detection effect on this data set.

**Table 3.** AP comparison of various algorithms on various types of datasets    %

| algorithm | normal | Dotted | Malposed | Unfiltered | Folded | mAP |
|---|---|---|---|---|---|---|
| Fast R-CNN | 70.22 | 76.98 | 60.71 | 81.20 | 69.99 | 71.82 |
| Faster R-CNN | 79.13 | 83.01 | 68.64 | 88.91 | 80.26 | 79.99 |
| SSD | 82.12 | 84.33 | 77.29 | 90.89 | 89.87 | 84.90 |
| YOLOv4 | 84.83 | 90.78 | 80.12 | 93.21 | 93.11 | 88.41 |
| YOLOv5 | 87.23 | 94.53 | 79.67 | 94.15 | 98.07 | 90.73 |
| CenterNet | 79.96 | 93.21 | 82.50 | 92.77 | 95.91 | 88.87 |
| C-CenterNet | **89.25** | **97.61** | **92.01** | **96.31** | **99.88** | **95.01** |

Figure 13 shows the visual detection and comparison results of some images. The detection results of Figure 13 show that when the images are clear and obvious, the five comparison methods can detect defects, but the detection confidence is different. The Faster R-CNN and SSD methods detect with low confidence and are indistinguishable for cases where one defect may represent two defect types. YOLOv5 has high overall detection confidence, but there are missed defects and a low recall rate. CenterNet is able to distinguish between cases where a defect may represent two defect types, but has low detection confidence. The method C-CenterNet can distinguish between two defect types, reduce the occurrence of missed detection, and achieve good confidence. Therefore, as can be seen from Figure 13, the present method yields better results on the cigarette appearance data set.

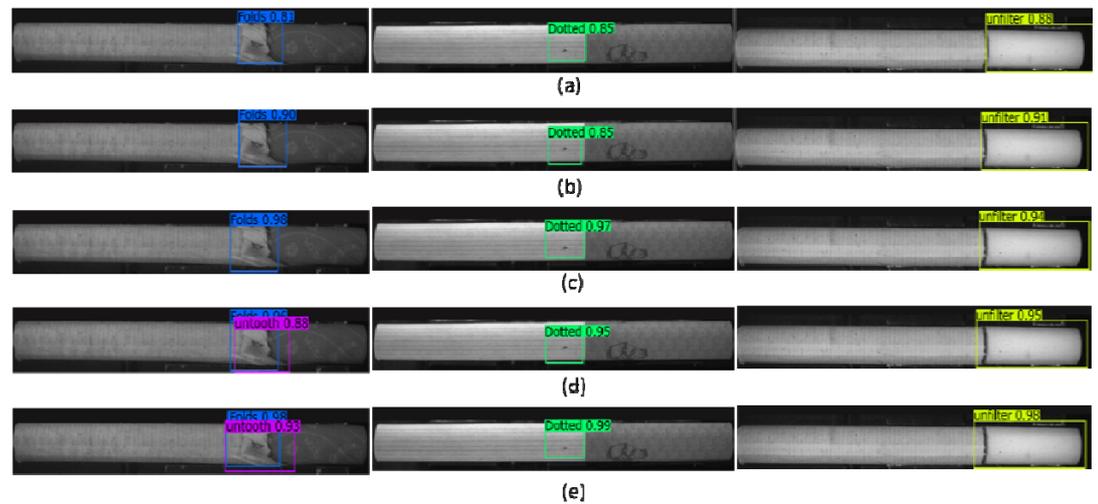

**Figure 13.** Comparison of the different network detection results: (a). Faster R-CNN; (b). SSD; (c). YOLOv5; (d). CenterNet; (e). C-CenterNet

*3.6 Improvement before and after comparison*

In order to verify the effectiveness of this method in terms of the accuracy improvement, the following experiments were conducted with the addition of a convolutional block attention mechanism, feature pyramid network, deformable convolution, and ACON activation function. The same parameter settings were utilized during the experiment to train the network. Table 4 shows the comparison of the accuracy, recall, and mAP values of the above modules. According to the table, the precision rate, recall rate, and mAP all increased to different degrees after addition. Data enhancement makes model images richer, model training is better, and the CBAM attention mechanism improves the precision; FPN further strengthens the feature fusion ability, and improves the precision and recall; DCN reduces the occurrence of missed detection and recall increases significantly; the ACON activation function adaptively adjusts the activation function, which results in a small increase in each index. Therefore, adding these modules has a positive effect on improving the accuracy, recall rate, and average detection accuracy of the model. Although the average detection time overall decreased by 1.2ms / branch, this time was also within the range of our expected time requirements.

**Table 4.** Comparison of the results of adding different modules

| Experiment | Data Augmentation | CBAM | FPN | DCN | ACON | P /% | R /% | mAP /% | mSpeed /(ms/branch) |
|---|---|---|---|---|---|---|---|---|---|
| 1 | - | - | - | - | - | 90.33 | 76.94 | 86.20 | 7.7 |
| 2 | √ | - | - | - | - | 92.65 | 78.30 | 88.87 | 7.7 |
| 3 | √ | √ | - | - | - | 96.91 | 80.02 | 91.52 | 8.3 |
| 4 | √ | √ | √ | - | - | 98.75 | 81.97 | 93.10 | 8.8 |
| 5 | √ | √ | √ | √ | - | 99.01 | 84.89 | 94.67 | 8.8 |
| 6 | √ | √ | √ | √ | √ | 99.89 | 85.96 | 95.01 | 8.9 |

Table 5 shows the comparison of AP changes for the five defect types in this paper. CenterNet with Resnet50 as the main backbone feature extraction network on the original cigarette appearance defect data set yielded 86.20% mAP. After a series of data enhancements, the mAP increased to 88.87%. Then, the convolutional block attention mechanism was introduced based on CenterNet, which led to a 2.45% rise in mAP. Due to the large amount of Sigmoid calculation in the convolutional attention mechanism, the Sigmoid activation function in the convolutional attention mechanism was then replaced with the h-Sigmoid activation function, which accelerated the model convergence and the mAP increased by 0.2%. In the process of cigarette appearance defect detection, there are a few defect information elements of small and medium image targets, which are easy to lose during down-sampling; we integrated the feature map FPN for reasonable feature fusion, and this method helped the model to obtain a 1.58% mAP increase. Later, in order to increase the identification effect of irregular features and improve the detection rate of incorrect teeth and fold defects, a deformable convolutional network was introduced. mAP increased by 1.57%, mainly reflected in the detection accuracy of malposed and folded defects. Finally, we replaced the ReLU activation function with the adaptive activation function ACON, and the model's mAP further increased by 0.34%.

**Table 5.** AP comparison of adding different modules                            %

| Experiment | Data Augmentation | CBAM | FPN | DCN | ACON | Normal | Dotted | Malposed | Unfiltered | Folded | mAP |
|---|---|---|---|---|---|---|---|---|---|---|---|
| 1 | - | - | - | - | - | 78.89 | 91.86 | 80.52 | 89.53 | 90.20 | 86.20 |
| 2 | √ | - | - | - | - | 79.96 | 93.21 | 82.50 | 92.77 | 95.91 | 88.87 |
| 3 | √ | √ | - | - | - | 82.29 | 95.27 | 85.98 | 95.20 | 98.88 | 91.52 |
| 4 | √ | √ | √ | - | - | 84.99 | 96.78 | 89.33 | 95.50 | 98.92 | 93.10 |
| 5 | √ | √ | √ | √ | - | 88.97 | 97.15 | 91.54 | 96.00 | 99.67 | 94.67 |
| 6 | √ | √ | √ | √ | √ | 89.25 | 97.61 | 92.01 | 96.31 | 99.88 | 95.01 |

The above results show that the improved method achieves a strong accuracy improvement. After C-CenterNet training was completed, we submitted an image for detection, and part of the effect is shown below in Figure 14. In the case of good detection conditions, blurred shooting, too strong light, and complex background, the algorithm can accurately detect cigarette defects, achieve a high degree of confidence, and the prediction frame is more suitable for the real position. This shows that the model is robust.

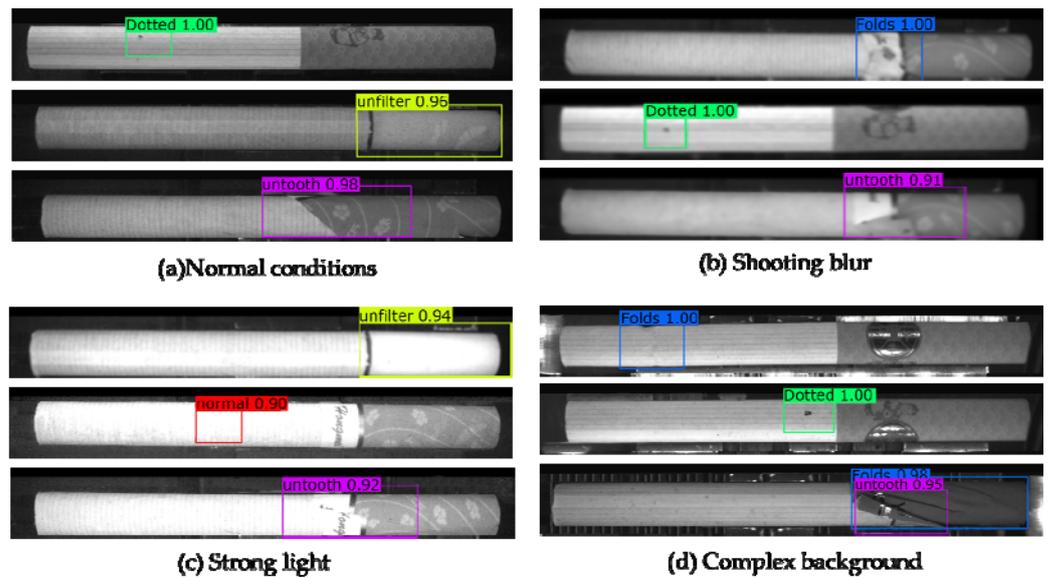

**Figure 14**. Example of experimental results

## 4. Conclusions

According to the characteristics of cigarette appearance defects, an improved CenterNet method, C-CenterNet, was used to detect cigarette appearance defects, and it could achieve good detection results in the case of an insufficient data set. The proposed method directly uses keypoint estimation to determine the defect center, class, and size of the keypoint. We selected the relatively lightweight Resnet50 as the backbone feature extraction network, and we added the convolutional attention mechanism, replacing part of the convolutional structure with deformable convolution. Then, we combined this system with the FPN to further strengthen the feature fusion, and we used the ACON activation function to adaptively select neuron activation or not. The experimental results demonstrate the effectiveness of the algorithm in the task of cigarette appearance defect detection. This method can achieve better detection precision and recall, and has certain robustness. The model achieves a large increase in accuracy with small changes in detection speed.

However, there are still many unreasonable places in the current C-CenterNet model. For example, in terms of detection speed, the performance of C-CenterNet needs to be further improved, and the current detection speed is slower than YOLOv5. The in-compatibility of speed and precision is still an unsolved problem. Although the improvement adopted in this paper significantly improves the detection accuracy, it also greatly increases parameters and computation time. In the future, we will focus on how to improve the speed of network detection, such as simplifying the network structure and speeding up the inference process. We hope that the next step can achieve better detection results in the detection of cigarette appearance defects, and deploy it into the embedded equipment of cigarette factories to complete the detection of cigarette appearance defects.

**Author Contributions:** H.L. proposed the network architecture design and the framework of the appearance defect detection method. L.Y. and K.L. collected and preprocessed the data set. L.Y. and H.L. performed the experiments. L.Y., H.L. and K.L. analyzed and discussed the experimental data. L.Y. wrote the paper. G.Y. and H.Z. revised the paper and provided valuable advice for the experiments. All authors have read and agreed to the published version of manuscript.

**Funding:** This research was funded by the Natural Science Foundation of China (Grant No. 62061049), the Application and Foundation Project of Yunnan Province (Grant No.202001BB050032), the Yunnan Provincial Department of Science and Technology - Yunnan

University Joint Special Project for Double-Class Construction, and the Open Project of CAS Key Laboratory of Solar Activity, National Astronomical Observatories (Grant No. KLSA202115).

**Acknowledgments:** We would like to thank the anonymous reviewers and the editor-in-chief for their comments that improved the paper. Thanks also to the data sharer. We thank all the individuals involved in the study.

**Conflicts of Interest:** The authors declare no conflicts of interest.